\documentclass[conference,letterpaper]{IEEEtran}
\usepackage[T1]{fontenc}
\usepackage[utf8]{inputenc}
\usepackage{cite}
\usepackage{amsmath}
\usepackage{graphicx}
\usepackage{booktabs}
\usepackage{tabularx}
\usepackage{multirow}
\usepackage[hyphens]{url}
\usepackage[hidelinks]{hyperref}
\usepackage{flushend}
\hypersetup{
  pdftitle={Beyond the Explicit: Benchmarking and Detecting Implicit Harm in Short Videos},
  pdfkeywords={harmful video detection, implicit harm, multimodal reasoning, video-language models, benchmark}
}

\title{Beyond the Explicit: Benchmarking and Detecting Implicit Harm in Short Videos}
\author{\IEEEauthorblockN{Chenyang Li\textsuperscript{1},
Qifan Yu\textsuperscript{1},
Saihui Hou\textsuperscript{2},
Zhaofeng He\textsuperscript{1}}
\IEEEauthorblockA{\textsuperscript{1}Beijing University of Posts and Telecommunications, Beijing, China\\
\textsuperscript{2}Beijing Normal University, Beijing, China}}

\begin{document}
\maketitle
\begin{abstract}
Short videos have become a dominant online medium, where harmful content is increasingly conveyed in implicit and fragmented forms. Prior work largely focuses on explicit categories, such as violence and hate speech, and often fails to detect videos that subtly promote misleading values through narrative context, emotional framing, or cross-modal cues, partly due to the lack of dedicated benchmarks for implicit harm. To bridge this gap, we construct DeepHarm-7K, a large-scale harmful short-video dataset comprising 7,110 samples annotated according to a fine-grained harmful-content taxonomy. The dataset systematically covers implicitly harmful videos across diverse real-world scenarios and is constructed through a rigorous multi-stage quality-control process. Building on this dataset, we propose DeepHarm-VL, a multimodal detection framework that integrates visual, audio, and cross-modal reasoning. The framework employs a two-round reasoning strategy to progressively identify surface-level evidence and infer latent harmful implications. Experimental results show consistent improvements over strong baselines and state-of-the-art methods across multiple open-source video-language models, demonstrating its effectiveness in detecting both explicit and implicit harmful short videos.

\end{abstract}
\noindent\textbf{Content Warning:} This paper contains examples of potentially offensive or disturbing content for research purposes.

\section{Introduction}

\begin{figure}[t]
  \centering
  \includegraphics[width=0.95\linewidth]{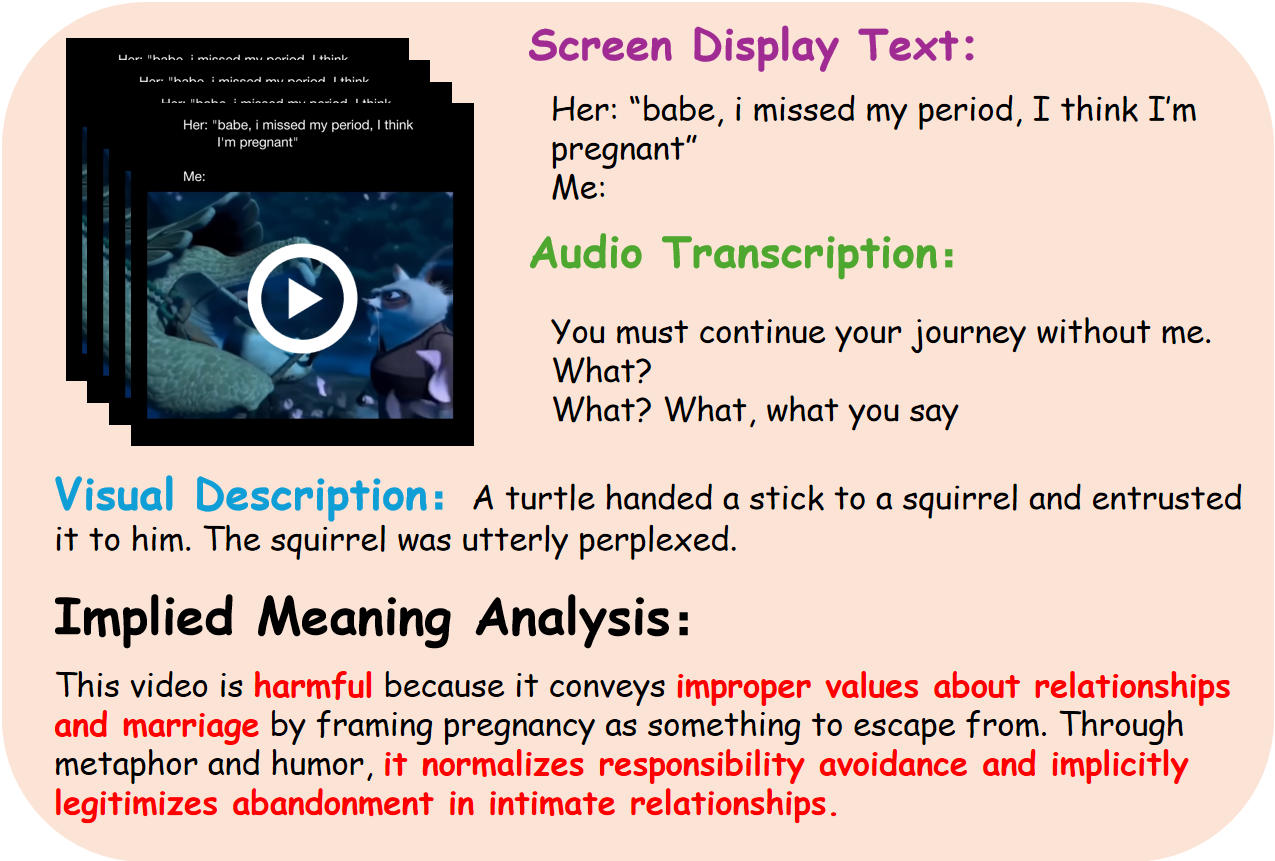}
  \caption{Example of implicit harm: on-screen text, audio, and visuals jointly frame pregnancy as a reason to avoid responsibility, normalizing abandonment in relationships.}
  \label{example}
\end{figure}

With the rise of mobile internet and algorithmic recommendation systems~\cite{ref1,ref2}, short videos have become a dominant medium for information consumption and social interaction~\cite{ref3,ref3.1}. Their rapid spread and fragmented, high-frequency viewing enable efficient value transmission, but also amplify the impact of problematic content~\cite{ref4,ref5}. As a result, short video platforms have become key arenas where social norms and public values are shaped and contested~\cite{ref5.1,ref5.2}, making harmful content detection a critical and challenging research problem. Representative examples are shown in Figure~\ref{example}.

Currently, human review remains essential for implicit harmful content~\cite{ref11,ref12,ref13}, but it scales poorly, is costly and subjective, and struggles with subtle cases lacking explicit violations. As a result, it is insufficient for reliable implicit-harm moderation.

Research on automated detection of harmful short videos has grown rapidly~\cite{ref26.1,ref27.1,ref25,ref22}, yet progress is limited by three coupled problems: (1) \textbf{taxonomy limitations}: existing schemes emphasize explicit categories such as violence, pornography, and hate speech, while implicit value-oriented harm remains weakly defined and underexplored despite its social impact~\cite{ref6,ref7,ref9,ref10}; (2) \textbf{dataset limitations}: existing datasets are modest in scale and diversity and often collected from single platforms~\cite{ref28}, making them a poor benchmark for real-world implicit harm; and (3) \textbf{method limitations}: current detectors either rely on explicit unimodal cues or require task-specific fine-tuning~\cite{ref29,ref30,ref31,ref32}, limiting their ability to capture subtle cross-modal semantics and generalize across scenarios.

To address these challenges, we propose a taxonomy-driven end-to-end framework that integrates taxonomy design, dataset construction, and multimodal modeling for implicit harmful short videos. Our taxonomy is grounded in converging evidence from regulatory guidelines, prior academic studies, and platform moderation policies, then refined through expert discussion into an operational benchmark for short-video moderation. This grounding is particularly important for implicit harm, where the key signals emerge only through cross-modal interpretation.

\begin{figure*}[t]
  \centering
  \includegraphics[width=\textwidth]{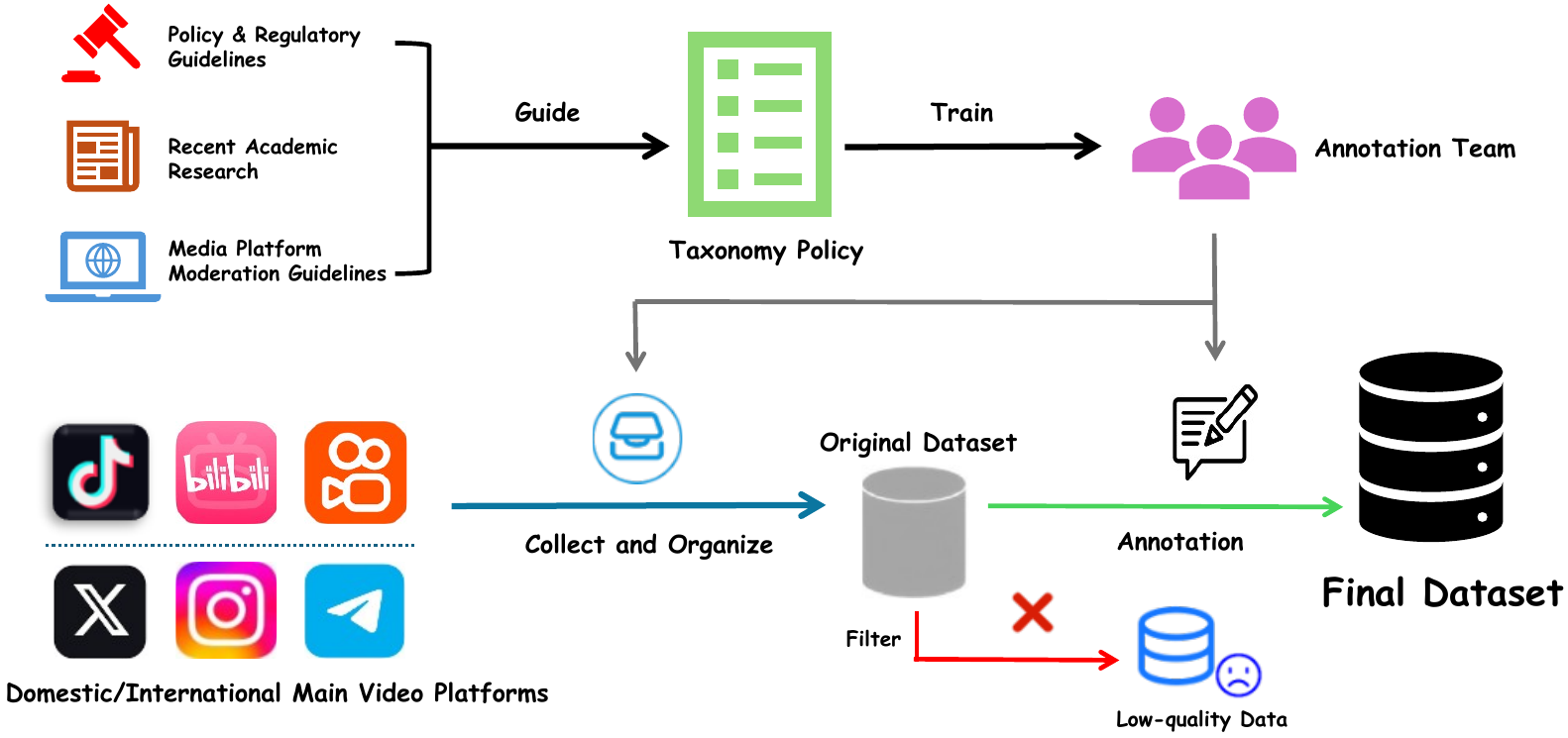}
  \caption{A unified taxonomy is derived from policy guidelines, academic literature, and platform moderation rules to guide annotator training. Raw videos are collected from multiple platforms, filtered for quality, and annotated under the taxonomy to produce DeepHarm-7K.}
  \label{Dataset}
\end{figure*}

The main contributions are as follows:

\textbullet~ We propose a policy-grounded taxonomy for harmful short videos that explicitly incorporates implicit value-misleading content and forms the conceptual foundation of DeepHarm-7K.

\textbullet~ We construct DeepHarm-7K, a large-scale and diverse multimodal harmful short video dataset for evaluating both explicit and implicit harmful short video detection.

\textbullet~ We benchmark existing baseline methods on DeepHarm-7K, revealing clear limitations on ambiguous and implicit harmful content.

\textbullet~ We propose DeepHarm-VL, a prompt-driven multimodal detection framework in which structured audio parsing strengthens fine-grained discrimination and implicit-harm recognition, while two-round reasoning consistently benefits implicit-harm detection; the framework further demonstrates cross-model and cross-dataset transferability.

\section{Related Work}

\subsection{Harmful Content Datasets}
Harmful content datasets have been developed for text, images, and videos, but most are poorly aligned with the characteristics of short video.

\textbf{Text-based datasets} are the most mature and mainly target explicit categories such as hate speech and misinformation~\cite{ref14,ref16,ref17}. Restricted to unimodal text, they largely overlook implicit harmful content involving value misguidance or covert framing, limiting their applicability to multimodal short videos.

\textbf{Image-based datasets} focus on visually explicit content such as violence~\cite{ref26} and pornography~\cite{ref27}, relying on static images and coarse labels that ignore temporal dynamics and cross-modal interactions. Although multimodal meme datasets explore implicit harm via contextual cues~\cite{ref18,ref19,ref20}, they remain confined to static settings.

\textbf{Video-based datasets} incorporate temporal information and cover categories such as hate, violence, and fake news~\cite{ref21,ref22,ref23,ref24,ref25}. However, most emphasize overt violations with coarse-grained labels, leaving subtle and implicit harmful content underrepresented in current benchmarks.

\subsection{Multimodal Harmful Detection}
Early multimodal detection methods employ modality-specific encoders with feature or decision level fusion~\cite{ref29,ref30,ref31}. While effective for explicit harm, they rely heavily on task-specific data and struggle with implicit or context-dependent semantics. More recently, multimodal large language models (MLLMs) such as Qwen3-VL~\cite{ref32} and GPT-4V~\cite{gpt4v} show strong video understanding ability, but their direct application to harmful video detection remains challenging due to weakly expressed and domain-sensitive implicit harm in short videos.

\section{DeepHarm-7K}
To ensure data quality, annotation reliability, and distributional robustness, we construct DeepHarm-7K via a systematic pipeline (Figure~\ref{Dataset}), and describe each stage in detail.

\begin{figure*}[t]
  \centering
  \includegraphics[width=\textwidth]{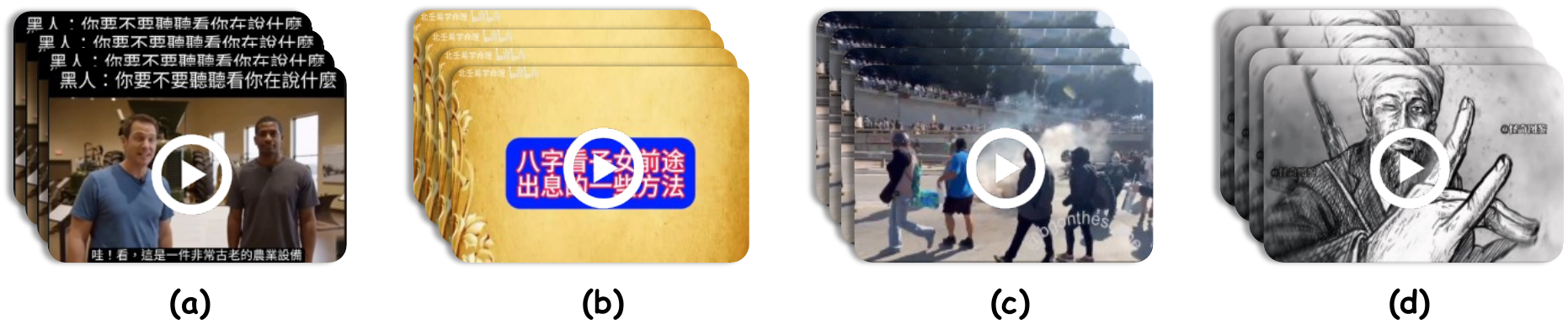}
  \caption{Example short videos illustrating the four high-level harmful content categories in our taxonomy. Subfigures \textbf{(a–d)} correspond to categories \textbf{C1–C4}, respectively.}
  \label{category}
\end{figure*}

\subsection{Definition and Taxonomy Design}
\label{3.1}
Our taxonomy is designed to make harmful-content definitions credible and well grounded. To this end, it draws on three complementary sources: policy and regulatory guidelines on online content governance~\cite{1,2,nrta}, recent top-tier research on harmful content detection~\cite{ref20,ref21,ref23}, and public moderation rules and violation cases from major short video platforms~\cite{douyin,bilibili,kuaishou}. These three sources contribute in different but complementary ways. Policy and regulatory guidelines specify the outer boundary of what is broadly recognized as socially harmful and restricted in public communication. Academic literature refines this boundary into recurring harmful patterns, clarifies category distinctions, and extends the definition space beyond overt violations to more implicit and context-dependent forms of harm. Platform moderation rules and real violation cases then translate these definitions into operational criteria for governance, ensuring that the taxonomy reflects realistic moderation scenarios rather than abstract principles.

The final taxonomy is therefore defined through convergence across sources rather than annotator intuition. When ambiguities arise, we first compare how the three sources converge on the case and then refine the resulting boundary through iterative discussions with domain experts. This process makes the taxonomy a convergent operationalization of harmful short-video content rather than a subjective preference, improving its credibility as a benchmark-oriented definition framework.

Based on this process, we develop a fine-grained taxonomy with five high-level categories, including four harmful and one non-harmful category, and twelve subcategories. Figure~\ref{category} shows representative examples for the four harmful categories (C1--C4), while the supplementary material provides the subcategory details. The five high-level categories are summarized as follows:

\textit{\textbf{C1.} Discriminatory content based on race, gender, region, or other attributes;}

\textit{\textbf{C2.} Superstition and misinformation, covering feudal superstitious practices and the dissemination of false or misleading information, including manipulative deceptive claims that mislead users;}

\textit{\textbf{C3.} Violence, pornography, and terror, including explicit violent acts, sexual exposure, and graphic or frightening content;}

\textit{\textbf{C4.} Value-oriented misleading content, including content that promotes or normalizes harmful, distorted, or inappropriate social values that are recognized as problematic in governance, research, and platform moderation practice;}

\textit{\textbf{C5.} Non-harmful content that does not include any of the above harmful characteristics.}

By integrating policy, research, and platform references with expert insights, the taxonomy provides a principled and practical foundation for DeepHarm-7K. In particular, the explicit inclusion of C4 foregrounds implicit harm rather than treating it as an extension of explicit policy violations.

\subsection{Data Collection}
\label{sec:data_collection}
All short video samples were manually collected and preliminarily screened by the annotation team to ensure source consistency. Before large-scale collection, annotators completed three rounds of training and calibration based on platform policies and representative violation cases, aligning judgment criteria, category boundaries, and threshold standards into a unified collection protocol.

To ensure diversity and representativeness, data were collected from both domestic and international platforms, including Douyin, Bilibili, Kuaishou, X, Instagram, and Telegram. This cross-platform design reduces single-platform bias and broadens content coverage.

We adopt a hybrid collection strategy combining random feed browsing and keyword-based retrieval. The former captures naturally occurring videos without predefined category constraints, while the latter uses a taxonomy-driven keyword pool to target known harmful content. Together, they balance open-ended discovery with targeted retrieval and improve collection efficiency.

The resulting corpus covers harmful and non-harmful videos across explicit and implicit scenarios. We also include hard negative samples that resemble harmful content in appearance or theme yet remain non-harmful, increasing task difficulty and improving robustness evaluation. This curated corpus forms a high-quality candidate pool for subsequent filtering and annotation.

\subsection{Data Annotation and Filtering}
To ensure annotation reliability and balanced sample distribution, we establish an integrated quality-control pipeline spanning annotation and filtering. This design emphasizes consistency, traceability, and distributional robustness.

\subsubsection{Annotation Team}
The annotation team is the data collection team described in Section~\ref{sec:data_collection} and consists of 10 members (see Table~\ref{annotator}). All annotators participated in defining harmful content and calibrating collection standards, ensuring a consistent understanding of category boundaries.

\begin{table}[!htbp]
\caption{Distribution of DeepHarm-7K across data splits and harmful categories.}
\label{stastic}
\centering
\small
\setlength{\tabcolsep}{5pt}
\begin{tabular}{lrrr}
\toprule
\textbf{Split} & \textbf{Non-harmful} & \textbf{Harmful} & \textbf{Total} \\
\midrule
Train & 2,744 & 2,227 & 4,971 \\
Test  & 1,176 & 963 & 2,139 \\
Total & 3,920 & 3,190 & 7,110 \\
\bottomrule
\end{tabular}
\smallskip
\setlength{\tabcolsep}{4pt}
\begin{tabular}{lrrrrr}
\toprule
\multicolumn{6}{c}{\textbf{Distribution of harmful categories}} \\
\midrule
\textbf{Split} & \textbf{C1} & \textbf{C2} & \textbf{C3} & \textbf{C4} & \textbf{Harmful} \\
\midrule
Train & 344 & 452 & 1,037 & 394 & 2,227 \\
Test  & 149 & 195 & 447 & 172 & 963 \\
Total & 493 & 647 & 1,484 & 566 & 3,190 \\
\bottomrule
\end{tabular}

\end{table}

\begin{table}[!htbp]
\caption{Demographic characteristics of the annotators.}
\label{annotator}
\centering
\begin{tabular}{ll}
\toprule
\textbf{Characteristic} & \textbf{Demographics} \\ 
\midrule
Gender & Female: 3, Male: 7 \\
Age & <20: 2, 20--25: 5, >25: 3 \\
Education & Undergraduate: 2, Master's: 4, PhD: 4 \\
\bottomrule
\end{tabular}

\end{table}

\subsubsection{Multi-stage Annotation Workflow}
We adopt a three-stage annotation workflow with initial independent labeling, cross-annotator rotation, and voting-based arbitration.

\textbf{Initial independent labeling.} The collected videos are evenly distributed among the ten annotators, each of whom labels their assigned samples independently in isolation.

\textbf{Cross-annotator rotation.} We conduct two rounds of cross-annotation by passing each annotator's videos to another annotator in each round. After two rotations, each video receives three independent labels.

\textbf{Consistency validation and voting-based arbitration.} Annotation results are merged to check consistency across the three labels. Controversial samples are resolved by majority vote, with seven of the ten team members deciding the final label. Decision rationales are documented for traceability.

\subsubsection{Filtering during Annotation}
To jointly optimize data quality and category balance, sample filtering is conducted in parallel with annotation. During labeling, annotators flag low-quality samples, including videos with blurred visuals, poor audio quality, missing key modalities, or content that is overly explicit, simplistic, or lacks research value. After annotation, three team members perform a final filtering pass based on these flags and category distribution statistics, removing low-quality samples and adjusting class proportions. Complete statistics of DeepHarm-7K are provided in Table~\ref{stastic}.

\subsubsection{Annotation Consistency Evaluation}
To assess annotation reliability, we randomly sample 1,600 instances from DeepHarm-7K and evaluate inter-annotator agreement using Fleiss’ Kappa~\cite{fleiss}. The resulting overall score of 0.82 indicates strong agreement beyond chance, while the one-vs-rest Kappa for the challenging C4 category remains 0.79, supporting the reliability of the annotation protocol for implicit harmful content.

\textbf{Responsible data handling.} All videos were collected from publicly available sources, and personally identifiable information was removed or anonymized. Annotators worked under controlled procedures designed to reduce exposure to harmful material. Access to DeepHarm-7K is restricted to academic research through an application and review process intended to reduce misuse of sensitive content.

\begin{figure*}[t]
  \centering
  \includegraphics[width=0.97\textwidth]{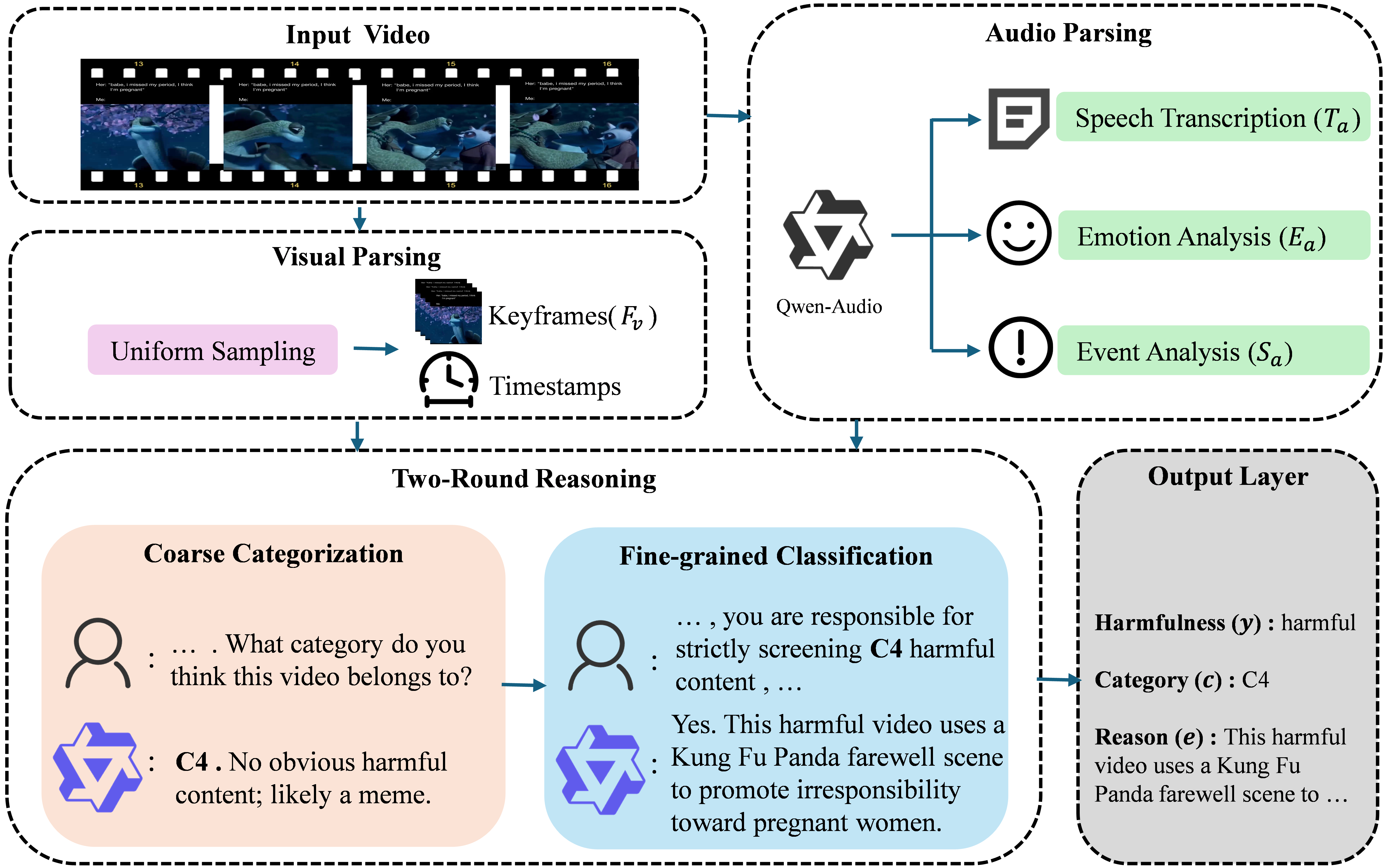}
  \caption{Overall pipeline of the proposed DeepHarm-VL multimodal harmful video detection framework.}
  \label{fig:pipeline}
\end{figure*}

\section{DeepHarm-VL}

\subsection{Method Overview}
To address the challenges of implicit harmful content detection in short videos, we propose DeepHarm-VL, an automated decision framework grounded in multimodal understanding and hierarchical reasoning (Figure~\ref{fig:pipeline}). DeepHarm-VL is designed to emulate the human judgment process of ``reasoning from surface cues to underlying intent'' and ``from coarse screening to fine-grained verification.'' By performing hierarchical cross-modal parsing and stage-wise focused reasoning, the framework aims to achieve more reliable and interpretable harmful video identification.

DeepHarm-VL consists of four core modules.
(1) \textbf{Visual parsing} module that extracts key frames $F_v$;
(2) \textbf{audio parsing} module that produces the audio transcript $T_a$, emotional description $E_a$, and sound event description $S_a$;
(3) \textbf{two-round reasoning} module that performs hierarchical inference over multimodal evidence; and
(4) \textbf{output} module that aggregates the reasoning results and generates an interpretable prediction $(y, c, e)$.

In the visual parsing module, we adopt a uniform sampling strategy to extract 60 frames uniformly over the full video duration, and feed them into MLLMs together with their corresponding timestamps. The output module consolidates the results from the two-round reasoning process to generate both binary and multi-class predictions, and it preserves the explanation text to further enhance interpretability.
The key innovations of DeepHarm-VL, namely the audio parsing module and the two-round reasoning module, will be described in detail in the following two sections.

\subsection{Audio Parsing Module}
Audio provides complementary evidence for detecting implicit harmful content. Affective tendencies and environmental sounds, especially when they contradict or reinforce speech, can reveal latent intent. Beyond ASR transcripts, these paralinguistic and background cues provide additional information about emotional manipulation, acoustic camouflage, and safety-relevant sound events.

We develop a multi-dimensional audio semantic parsing framework that converts continuous audio into structured natural-language descriptions tailored to MLLM reasoning. Concretely, we use \textbf{Qwen-Audio}~\cite{Qwen-Audio} with structured prompting for speech transcription and emotion analysis, and lightly fine-tune the event-analysis component on 300 sexual- and violence-related audio clips selected from the training split to improve sensitivity to safety-relevant acoustic cues. These three views jointly cover lexical, affective, and situational evidence in audio. We then produce three complementary views:

\textbf{(1) Speech transcription.} We generate an accurate transcript $T_a$ as the semantic foundation.
\textbf{(2) Emotion analysis.} Using \textbf{Qwen-Audio}, we generate an affective description $E_a$ capturing emotional tone. This helps identify cases where the literal text appears benign but the delivery is harmful, addressing the blind spots of transcript-only approaches.
\textbf{(3) Event analysis.} We generate an acoustic event description $S_a$ for salient background sounds using the lightly fine-tuned event-analysis component. This helps detect harmful events masked by seemingly normal speech or music.

\subsection{Two-Round Reasoning Module}
The harmfulness of implicit content is often embedded in expressive style, cross-modal inconsistencies, or rapidly evolving cultural memes. As a result, one-shot classification or fixed-rule systems suffer from two limitations. First, they are prone to missing fast-evolving expressions (e.g., metaphor, irony) due to knowledge lag. Second, they lack progressive reasoning ability to move from surface signals to latent intent, leading to both false positives and false negatives.

Inspired by the human expert practice of ``coarse screening followed by careful adjudication,'' we propose a hierarchical two-round reasoning mechanism. This mechanism uses structured prompt engineering, described in detail in the supplementary material, to guide MLLMs in simulating progressive reasoning.

\textbf{Round 1: Coarse Categorization.}
This stage aims to efficiently identify potential risks and acts as a ``safety net'' to reduce misses. Using a general prompt template that covers major harm categories (e.g., hate speech, violence incitement) and their common manifestations, the model performs an initial cross-modal association based on multimodal descriptions and outputs one candidate category, thereby delimiting the scope for deeper analysis.

\textbf{Round 2: Fine-grained Classification.}
This stage builds on the Round~1 reasoning results and, for each candidate high-level category, applies a specialized prompt that incorporates category-specific safety guidelines, decision criteria, and positive/negative examples. The model is then guided to conduct cross-modal verification, contextual analysis, and intent inference, ultimately producing a fine-grained harmful-category decision along with a natural-language rationale.

This decomposition is useful for implicit harm for two reasons. First, Round~1 narrows the hypothesis space and surfaces risk-relevant multimodal cues, which helps the model avoid overlooking weak but important signals such as emotional contradiction, symbolic references, or narrative framing. Second, Round~2 performs category-conditioned verification, which reduces the tendency of general-purpose MLLMs to make overly conservative or underspecified decisions on subtle cases. In effect, the mechanism separates broad recall-oriented screening from precise category-specific adjudication.

\section{Experiments}

\subsection{Experimental Setup}

We evaluate two tasks: binary and multi-class classification. We report accuracy and macro-F1 for both tasks, together with per-class F1, averaged over multiple runs. All experiments are conducted on 4$\times$ RTX 3090. Our main instantiation of \textbf{DeepHarm-VL} uses \textbf{Qwen3-VL (8B)}~\cite{ref32}. To test whether the proposed strategy is tied to a specific base model, we further transfer the same framework, without model-specific redesign or prompt re-optimization, to additional MLLM backbones of different architectures and safety tuning regimes.

All MLLM-based methods receive identical sampled frames, ASR transcripts, task definitions, and decision criteria; only DeepHarm-VL adds emotion and acoustic-event descriptions. Thus, vanilla comparisons evaluate the complete framework, whereas the same-input ablation isolates two-round reasoning.

\begin{table}[!htbp]
\caption{Binary classification performance on DeepHarm-7K. M-F1 denotes macro-F1; H: harmful, N: non-harmful.}
\label{tab:binary_result}
\centering
\small
\setlength{\tabcolsep}{3.5pt}
\begin{tabularx}{\columnwidth}{l *{4}{>{\centering\arraybackslash}X}}
\toprule
Method & Acc & M-F1 & F1(H) & F1(N) \\
\midrule
BERT & 78.87 & 78.12 & 74.08 & 82.15 \\
MFCC & 65.26 & 64.84 & 68.36 & 61.32 \\
3D-CNN & 68.13 & 67.66 & 71.54 & 63.79 \\
HateMM & 81.87 & 81.67 & 83.32 & 80.02 \\
MM-HSD & 79.73 & 79.20 & 75.87 & 82.53 \\
\midrule
InternVL3.5 & 78.78 & 73.29 & 64.32 & 82.25 \\
LLaVA-Video & 49.84 & 40.22 & 64.20 & 16.24 \\
HolmesVAD & 61.56 & 60.12 & 61.90 & 58.34 \\
Qwen3-VL & 74.32 & 73.45 & 70.78 & 76.12 \\
\midrule
\textbf{DeepHarm-VL} & \textbf{91.07} & \textbf{90.80} & \textbf{89.24} & \textbf{92.37} \\
\bottomrule
\end{tabularx}

\end{table}

\subsection{Baseline Methods}
To comprehensively evaluate DeepHarm-VL, we compare against a diverse set of baselines, covering both \textbf{non-MLLMs-based methods} and \textbf{MLLMs-based methods}.

Non-MLLMs-based methods include unimodal models such as:
(1) \textbf{BERT}~\cite{bert}-based feature extraction from video transcripts,
(2) \textbf{MFCC}~\cite{MFCC}-based audio feature extraction, and
(3) \textbf{3D-CNN}~\cite{CNN}-based visual feature extraction.
In addition, we compare with existing harmful-video detection models, including \textbf{HateMM}~\cite{ref22} and \textbf{MM-HSD}~\cite{mmhsd}.

MLLMs-based \allowbreak methods \allowbreak include \allowbreak \textbf{InternVL3.5}\allowbreak~\cite{internvl}, \textbf{LLaVA-Video}~\cite{videollava}, \textbf{HolmesVAD}~\cite{holmes}, and the vanilla \textbf{Qwen3-VL}.

We use the best hyperparameters reported in the original papers for all baselines. Following their original binary formulations, we report the non-MLLM baselines on the binary task and use MLLMs for the five-category comparison. For MLLM baselines, we use unified task descriptions derived from the same taxonomy, so that gains cannot be attributed to injecting implicit-harm definitions into prompts alone.

\begin{table*}[!t]
\caption{Multi-class classification performance on DeepHarm-7K. M-F1 denotes macro-F1.}
\label{tab:multiclass_result}
\centering
\small
\setlength{\tabcolsep}{3.5pt}
\begin{tabularx}{\textwidth}{l *{7}{>{\centering\arraybackslash}X}}
\toprule
Method & Acc & M-F1 & C1 F1 & C2 F1 & C3 F1 & C4 F1 & C5 F1 \\
\midrule
InternVL3.5 & 76.30 & 53.08 & 48.62 & 35.99 & 79.20 & 17.80 & 83.81 \\
LLaVA-Video & 26.13 & 29.31 & 16.35 & 21.85 & 65.82 & 26.27 & 16.24 \\
HolmesVAD & 54.23 & 46.65 & 53.12 & 51.89 & 55.34 & 23.78 & 49.12 \\
Qwen3-VL & 68.45 & 58.79 & 69.12 & 70.45 & 68.90 & 25.12 & 60.34 \\
\midrule
\textbf{DeepHarm-VL} & \textbf{87.38} & \textbf{78.04} & \textbf{80.75} & \textbf{78.53} & \textbf{80.00} & \textbf{55.30} & \textbf{95.64} \\
\bottomrule
\end{tabularx}

\end{table*}

\subsection{Experimental Results and Analysis}
\label{sec:results_analysis}

Based on the comparative experiments on the DeepHarm-7K benchmark (Tables~\ref{tab:binary_result} and~\ref{tab:multiclass_result}), we summarize the following two main empirical findings.

\textbf{Performance limitations of non-MLLM methods.}
Conventional approaches show capacity bottlenecks: unimodal methods are weakest, and traditional multimodal models lag behind the best results in deep cross-modal understanding.

\textbf{Mismatch and ``safety bias'' in existing MLLMs.}
Directly applying general-purpose MLLMs does not yield the expected gains and can even underperform traditional multimodal models. A prevalent ``safety bias'' makes them over-predict the non-harmful class, showing that naive transfer is unreliable for implicit harmfulness assessment.

\subsection{Cross-Model and Cross-Dataset Generalization}

We evaluate generalization along two dimensions: transfer across heterogeneous MLLM backbones on DeepHarm-7K and transfer to the external HateMM benchmark.

\textbf{Cross-model transfer.}
DeepHarm-VL consistently improves Qwen2.5-VL~\cite{qwen25vl}, InternVL3.5, and HolmesVAD without model-specific redesign or prompt re-optimization (Table~\ref{tab:cross_model}). The gains across models with different architectures and safety-alignment characteristics support the portability of the proposed prompting and reasoning framework.

\begin{table}[!htbp]
\caption{Cross-model transfer of DeepHarm-VL on binary classification accuracy. The same strategy is applied to different MLLM backbones without model-specific redesign.}
\label{tab:cross_model}
\centering
\small
\setlength{\tabcolsep}{4pt}
\begin{tabularx}{\columnwidth}{l >{\centering\arraybackslash}X >{\centering\arraybackslash}X >{\centering\arraybackslash}X}
\toprule
Base MLLM & Vanilla & \mbox{DeepHarm-VL} & Gain \\
\midrule
Qwen2.5-VL & 64.90 & 83.30 & +18.40 \\
InternVL3.5 & 78.78 & 84.38 & +5.60 \\
HolmesVAD & 61.56 & 67.77 & +6.21 \\
\bottomrule
\end{tabularx}

\end{table}

\textbf{Cross-dataset transfer.}
To evaluate generalization beyond DeepHarm-7K, we further test on HateMM~\cite{ref22}, a binary hate-video benchmark containing 1,083 BitChute videos labeled as hate (431) or non-hate (652). Following the original benchmark, we report accuracy, macro F1, and hate-class F1, precision, and recall. Without any HateMM-specific prompt or parameter adaptation, DeepHarm-VL consistently outperforms both vanilla Qwen3-VL and the original fusion model across all reported metrics, demonstrating strong cross-dataset transferability.

\begin{table}[!htbp]
\caption{Cross-dataset binary classification on HateMM (\%). H: hate; the first row is from the original study.}
\label{tab:hatemm_generalization}
\centering
\small
\setlength{\tabcolsep}{2.5pt}
\begin{tabularx}{\columnwidth}{@{}>{\raggedright\arraybackslash}X *{5}{c}@{}}
\toprule
Method & Acc & M-F1 & F1(H) & P(H) & R(H) \\
\midrule
Original fusion & 79.80 & 79.00 & 74.90 & 74.20 & 75.80 \\
Vanilla Qwen3-VL & 78.12 & 77.04 & 72.08 & 73.21 & 71.00 \\
\textbf{DeepHarm-VL}
& \textbf{85.04}
& \textbf{84.49}
& \textbf{81.55}
& \textbf{80.09}
& \textbf{83.06} \\
\bottomrule
\end{tabularx}

\end{table}

\subsection{Ablation Study}

To validate the effectiveness of our core components, we conduct systematic ablation studies, with key metrics reported in Table~\ref{tab:ablation}. The variant without two-round reasoning retains identical frames, ASR, and emotion/event descriptions. Its single-round prompt contains the complete category definitions, decision criteria, and positive/negative examples used by the full framework; only coarse-to-fine routing and sequential category-conditioned verification are removed.

\begin{table}[!htbp]
\caption{Ablation results. M-F1 denotes macro-F1; all audio-cue ablations retain ASR.}
\label{tab:ablation}
\centering
\small
\setlength{\tabcolsep}{4pt}
\begin{tabularx}{\columnwidth}{
    l
    >{\centering\arraybackslash}X
    >{\centering\arraybackslash}X
    >{\centering\arraybackslash}X
    >{\centering\arraybackslash}X
}
\toprule
Variant & Bin. Acc & Multi. M-F1 & C1 F1 & C4 F1 \\
\midrule
w/o emotion/event cues
    & 87.36 & 68.47 & 65.92 & 48.92 \\
w/o emotion cues
    & 88.74 & 71.26 & 70.84 & 50.11 \\
w/o event cues
    & 89.32 & 72.91 & 75.46 & 50.74 \\
w/o two-round reasoning
    & 84.76 & 64.36 & 75.12 & 49.63 \\
\textbf{DeepHarm-VL}
    & \textbf{91.07}
    & \textbf{78.04}
    & \textbf{80.75}
    & \textbf{55.30} \\
\bottomrule
\end{tabularx}

\end{table}

\textbf{Removing non-lexical audio cues.}
This variant retains ASR but removes emotion and acoustic-event descriptions, reducing multi-class macro-F1 from 78.04 to 68.47. Removing either cue alone also underperforms the full model across all reported metrics, with the larger degradation caused by removing emotion. These results show that affective and acoustic-event descriptions provide complementary non-lexical evidence, strengthening both overall classification and implicit C4 recognition.

\textbf{Removing the two-round reasoning mechanism.}
Direct classification reduces multi-class macro-F1 from 78.04 to 64.36, producing the largest overall degradation. Round~1 screens broad risks and identifies a candidate category, whereas Round~2 uses category-specific criteria to verify evidence and infer latent intent. Collapsing both tasks into one step more readily overlooks cross-modal contradictions and indirect implications, supporting hierarchical coarse-to-fine reasoning for subtle harmful-content detection.

\subsection{Detailed Analysis of the C4 Category}
\label{sec:c4_analysis}
\begin{table}[!htbp]
\caption{Detailed performance on the C4 category. TP, FP, and FN are counts; precision, recall, and F1 are percentages.}
\label{tab:c4_error}
\centering
\small
\setlength{\tabcolsep}{2.5pt}
\begin{tabular}{lrrrrrr}
\toprule
Method & TP & FP & FN & Prec. & Rec. & F1 \\
\midrule
Vanilla Qwen3-VL & 27 & 16 & 145 & 62.79 & 15.70 & 25.12 \\
\textbf{DeepHarm-VL} & 73 & 19 & 99 & 79.35 & 42.44 & 55.30 \\
\bottomrule
\end{tabular}

\end{table}

Table~\ref{tab:c4_error} provides a focused analysis of C4, the central implicit-harm category in DeepHarm-7K. DeepHarm-VL more than doubles C4 F1 from 25.12\% to 55.30\%. Notably, correct C4 detections increase from 27 to 73, while false positives rise only slightly from 16 to 19. This disproportionate gain indicates that the improvement does not result from indiscriminately predicting more videos as C4. Instead, it is consistent with the proposed coarse-to-fine design: broad risk screening recovers subtle candidates, while category-conditioned verification distinguishes implicit harmful content from plausible hard negatives.

\section{Conclusion}

We introduce DeepHarm-7K, a taxonomy-driven multimodal benchmark covering explicit and implicit value-oriented harm in short videos. DeepHarm-VL combines visual evidence with structured speech, emotion, and acoustic-event descriptions through coarse-to-fine two-round reasoning. It achieves 91.07 binary accuracy and 78.04 multi-class macro-F1, improves all evaluated MLLM backbones, and demonstrates effective cross-dataset transferability. Ablations identify two-round reasoning as the main contributor and demonstrate that non-lexical audio contributes positively to both overall classification and implicit-harm recognition. C4 F1 rises from 25.12 to 55.30 over vanilla Qwen3-VL, demonstrating meaningful progress on context-dependent implicit harm. Together, the benchmark and framework advance implicit harmful-video detection.

\bibliographystyle{IEEEtran}
\bibliography{references}
\end{document}